\pdfoutput=1

\documentclass[11pt]{article}

\usepackage[preprint]{acl}

\usepackage{times}
\usepackage{latexsym}

\usepackage[T1]{fontenc}

\usepackage[utf8]{inputenc}

\usepackage{microtype}

\usepackage{inconsolata}

\usepackage{graphicx}

\usepackage{hyperref}
\usepackage{url}
\usepackage{comment}
\usepackage{amsmath}
\usepackage{booktabs}
\usepackage{multirow}
\usepackage{makecell}
\usepackage{longtable}
\usepackage{amssymb}

\title{PC-Agent: A Hierarchical Multi-Agent Collaboration Framework for Complex Task Automation on PC}

\author{
\textbf{Haowei Liu}$^{1,2*}$, \textbf{Xi Zhang}$^{3*}$, \textbf{Haiyang Xu}$^{3\dagger}$,
\textbf{Yuyang Wanyan}$^{1,2}$, \textbf{Junyang Wang}$^{4}$, \textbf{Ming Yan}$^{3\dagger}$, \\
\textbf{Ji Zhang}$^{3}$, \textbf{Chunfeng Yuan}$^{1,2\dagger}$,
\textbf{Changsheng Xu}$^{1,2}$, \textbf{Weiming Hu}$^{1,2,5}$, \textbf{Fei Huang}$^{3}$\\
$^{1}$MAIS, Institute of Automation, Chinese Academy of Sciences, China \\
$^{2}$School of Artificial Intelligence, University of Chinese Academy of Sciences, China \\
$^{3}$Alibaba Group
$^{4}$Beijing Jiaotong University \\
$^{5}$School of Information Science and Technology, ShanghaiTech University, China \\
\texttt{liuhaowei2019@ia.ac.cn}, \texttt{cfyuan@nlpr.ia.ac.cn} \\
\texttt{\{shuofeng.xhy, ym119608\}@alibaba-inc.com} \\
}

\begin{document}
\maketitle

\let\thefootnote\relax\footnotetext{$^*$The first two authors contributed equally to this work.}
\let\thefootnote\relax\footnotetext{$^\dagger$Corresponding authors.}

\begin{abstract}
In the field of MLLM-based GUI agents, compared to smartphones, the PC scenario not only features a more complex interactive environment, but also involves more intricate intra- and inter-app workflows. To address these issues, we propose a hierarchical agent framework named PC-Agent. Specifically, from the perception perspective, we devise an Active Perception Module (APM) to overcome the inadequate abilities of current MLLMs in perceiving screenshot content. From the decision-making perspective, to handle complex user instructions and interdependent subtasks more effectively, we propose a hierarchical multi-agent collaboration architecture that decomposes decision-making processes into Instruction-Subtask-Action levels. Within this architecture, three agents (\textit{i.e.}, Manager, Progress and Decision) are set up for instruction decomposition, progress tracking and step-by-step decision-making respectively. Additionally, a Reflection agent is adopted to enable timely bottom-up error feedback and adjustment. We also introduce a new benchmark PC-Eval with 25 real-world complex instructions. Empirical results on PC-Eval show that our PC-Agent achieves a 32\% absolute improvement of task success rate over previous state-of-the-art methods. The code is available at \url{https://github.com/X-PLUG/MobileAgent/tree/main/PC-Agent}.
\end{abstract}

\section{Introduction}

\begin{figure*}[t]
    \centering
    \includegraphics[width=1.0\textwidth]{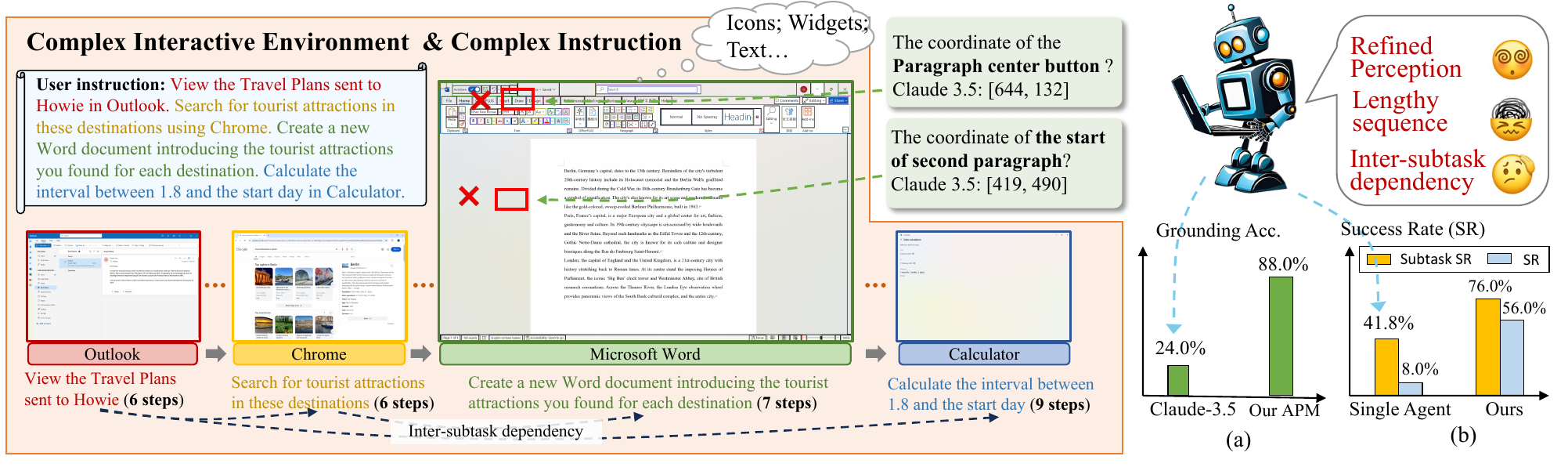}
    \vspace{-6mm}
    \caption{Illustration of the complexity of the PC scenario: (1) Complex interactive environment with dense and diverse elements. (2) Long and complex task sequences containing intra- and inter-software workflows. 
    }
    \label{fig1}
\vspace{-4mm}
\end{figure*}

Recently, Multi-modal Large Language Models (MLLM)~\cite{bai2023qwen,ye2024mplug,chen2024internvl,li2024llava} have achieved impressive progress across various domains.
Building on the powerful perception and reasoning abilities of MLLMs, researchers have extended them into multi-modal agents to assist humans in completing various tasks. In this field, graphical user interface (GUI) agents have garnered significant attention~\cite{wang2024mobilev1, agashe2024agent, zhang2023appagent, wang2024oscar}, as the automation of smart devices (\textit{e.g.}, smartphones, PCs) by agents holds vast application potential.

Compared to smartphones, the complexity of the PC scenario manifests in two aspects: 
\textbf{(1) More complex interactive environment.}
PC's GUI encompasses denser and more diverse interactive elements (\textit{i.e.}, icons and widgets),
along with varied text layouts (\textit{e.g.}, documents in \textit{Word} and code in \textit{VS Code}), posing significant challenges for screen perception.
For example, as Figure~\ref{fig1} shows, the top ribbon of \textit{Word} contains a plethora of icons and widgets, yet lacks textual labels indicating their functions.
As a result, even state-of-the-art MLLMs (\textit{e.g.}, Claude-3.5) exhibit inadequate abilities in perceiving and grounding icons and text on PC screens,
{and only achieves 24.0\% accuracy on a GUI grounding dataset~\footnote{More details about the grounding dataset can be found in Appendix~\ref{app_3}} in Figure~\ref{fig1}(a)}.
\textbf{(2) More complex task sequences.}
Compared to smartphones, PCs are generally used in productivity scenarios that involve more intricate intra- and inter-app workflows, and require longer and more intricate operation steps.
Taking \textit{making a travel plan} on PC (as in Figure~\ref{fig1}) as an example, it might involve multiple subtasks across four applications.
As a result,
on the one hand, the lengthy operation sequences (\textit{i.e.}, 28 steps in total) increase the difficulty of sensing the task progress.
On the other hand, 
the existence of inter-subtask dependencies 
requires the agent to consider the execution results of preceding subtasks when making decisions,
further increasing the decision-making difficulty.
As Figure \ref{fig1}(b) shows, the instruction-level success rate (SR) of a single agent (GPT-4o~\citealp{hurst2024gpt}) declines drastically from 41.8\% to 8\% compared to subtask SR,
highlighting the challenge of completing real-world instructions on PC.
To handle cross-app tasks, the previous work UFO~\cite{ufo} designs a dual-agent framework,
one for application selection and the other for specific control interactions.
To tackle complex PC tasks,
Agent-S~\cite{agashe2024agent} combines online search and local memory for experience-augmented planning.
However, these methods lack fine-grained perception and operation abilities of on-screen text, which is crucial in productivity scenarios (\textit{e.g.}, Word document editing).
Moreover, they generally overlook the complex dependencies between subtasks, thereby exhibiting limited abilities on realistic intra- and inter-app complex tasks.

\begin{figure*}[t]
    \centering
    \includegraphics[width=0.95\textwidth]{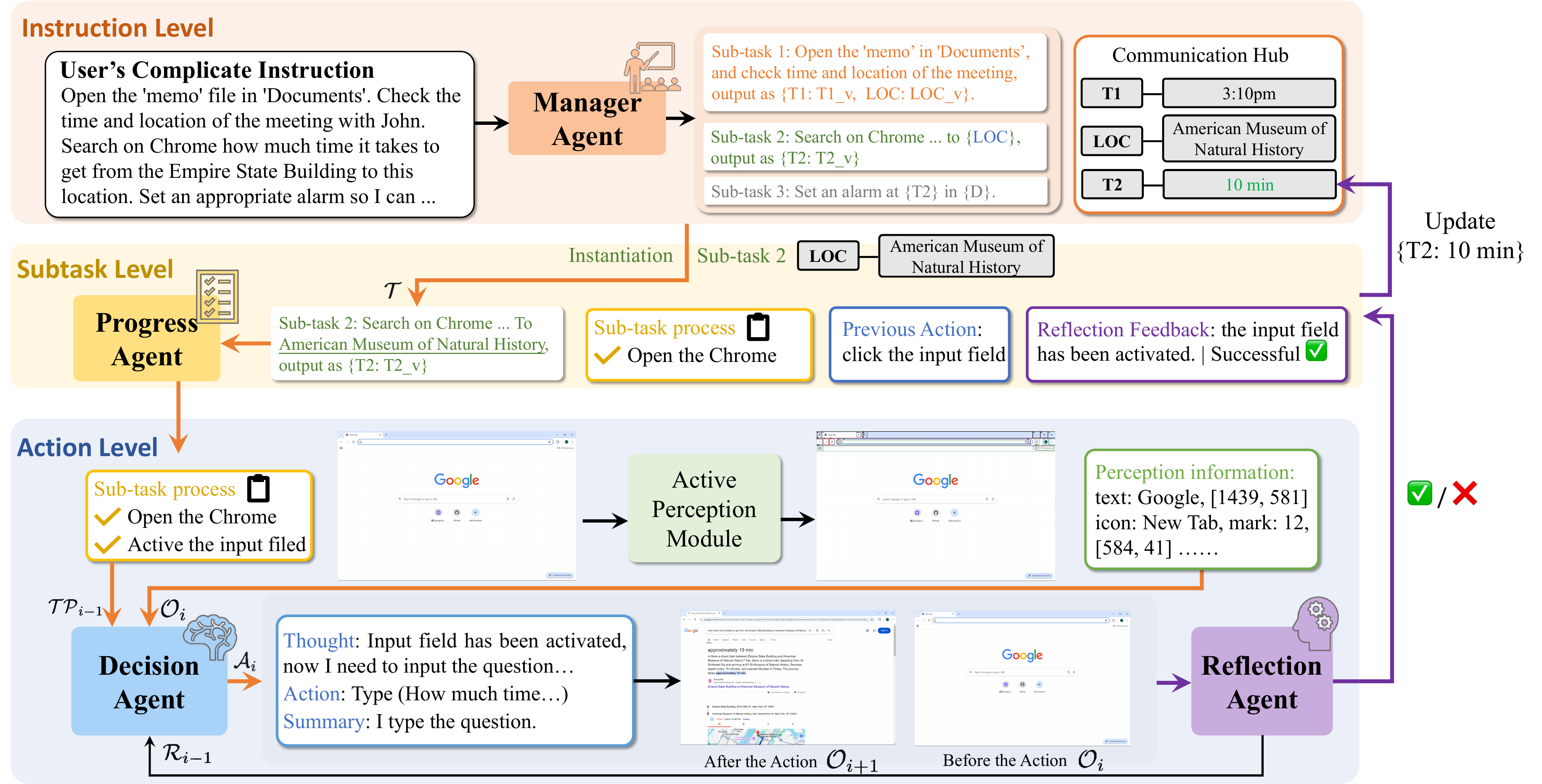}
    \caption{\small Overview of the proposed PC-Agent, which decomposes the decision-making process into three levels.
    The orange lines denote the top-down decision-making decomposition, and the purple lines represent the bottom-up reflection process.
    }
    \label{fig2}
    \vspace{-2mm}
\end{figure*}

In this paper, we propose the PC-Agent framework to handle the complex interactive environment and complex tasks in PC scenarios, which comprises three core designs:
\textbf{(1) Active Perception Module.} 
To enhance the fine-grained perception and operation abilities of the agent, we propose an Active Perception Module (APM). For interactive elements, we use the accessibility tree to extract their locations and meanings. For text, we employ an MLLM-driven intention understanding agent for target text extraction, followed by OCR to obtain precise locations.
\textbf{(2) Hierarchical Multi-agent Collaboration.} 
To improve the abilities of handling complex instructions, we adopt a divide-and-conquer approach and propose a Hierarchical Multi-agent Collaboration architecture.
Specifically, we break down decision processes into three levels: \textbf{Instruction-Subtask-Action.}
At the instruction level, a Manager Agent (MA) decomposes the user instruction into parameterized subtasks, with significantly fewer operation steps and lower decision-making difficulty.
The MA also manages inter-subtask communication to handle complex dependencies between them.
At the subtask level, a Progress Agent (PA) tracks and summarizes operation history for precise progress awareness.
At the action level, a Decision Agent (DA) makes decisions step-by-step by combining the APM's perception information and PA's progress information, and interacts with the PC environment to complete the decomposed subtasks.
\textbf{(3) Reflection-based Dynamic Decision-making.}
Building on the above
architecture, we also introduce a reflection-based dynamic decision-making mechanism for error detection in execution results, with timely feedback and adjustments. An additional Reflection Agent (RA) is set at the action level to observe screen changes before and after DA decisions, assessing the correctness of this step and conveying feedback to the DA and PA.
Figure~\ref{fig2} shows the entire process.
Combining the hierarchical multi-agent collaboration architecture with reflection-based dynamic decision-making,
our PC-Agent framework can \textbf{decompose complex user instructions from top to bottom} and \textbf{provide precise feedback from bottom to top during execution}.
Consequently, the four agents collaborate to alleviate the difficulty of interactive environments and complex workflow tasks on PC.

To better evaluate the capabilities of agents on complex tasks, we present a new benchmark \textbf{PC-Eval} for PC productivity environments.
PC-Eval comprises 8 popular applications and 25 complex user instructions, each consisting of several interdependent subtasks.
It provides a challenging and realistic benchmark, by emphasizing complex workflows and long-horizon decision-making.
Comparing our PC-Agent with advanced MLLM-based single agents and existing open-source PC agents on PC-Eval, we can find that PC-Agent achieves significant improvements in both instruction- and subtask-level success rates, demonstrating the effectiveness of the proposed framework.

Our contributions can be summarized as follows:

\noindent \textbf{(1)} We propose a PC-Agent framework to overcome the limitations of existing methods in handling complex interactive environments and complex tasks in PC scenarios.
An Active Perception Module (APM) is devised to enable PC-Agent with refined perception and operation capabilities.

\noindent \textbf{(2)} To tackle complex PC tasks, we propose a hierarchical multi-agent collaboration architecture decomposing the decision process into three levels (\textit{i.e.}, instruction-subtask-action),
and introduce a reflection-based dynamic decision-making mechanism for timely error feedback and adjustments.

\noindent \textbf{(3)} We create a PC-Eval benchmark involving 8 commonly used PC applications
to better assess the agent's capabilities in handling complex user instructions. Experimental results demonstrate that the proposed PC-Agent largely outperforms previous methods in completing complex PC tasks.

\section{PC-Agent}

\subsection{Task Formulation}
Given an GUI environment and a user instruction \(\mathcal{I}\), the GUI Agent (denoted as \(\rho\)) obtains an observation \(\mathcal{O}\) (\textit{e.g.}, screenshot) about 
the environment. Based on internal reasoning and planning, it makes a decision about the current step's action \(\mathcal{A}\),
which interacts with the GUI environment and alters the environment's state. 
This process generally occurs step-by-step. 
This process can be formalized as:

\begin{equation}
\mathcal{A}_i = \rho(\mathcal{I}, \mathcal{O}_i, \mathcal{H}_{i-1}),
\label{equ:formulation}
\end{equation}
where $\mathcal{A}_i$ and $\mathcal{O}_i$ represent the action and observation in the $i$-th step, and $\mathcal{H}_{i-1}$ is the operation history until the $(i-1)$-th step.
The complex interactive environments and task sequences in PC increases the complexity of \(\mathcal{I}\), \(\mathcal{O}\) and \(\mathcal{H}\), necessitating designing an agent framework tailored for complex PC scenarios.

\begin{figure*}[t]
    \centering
    \includegraphics[width=0.95\textwidth]{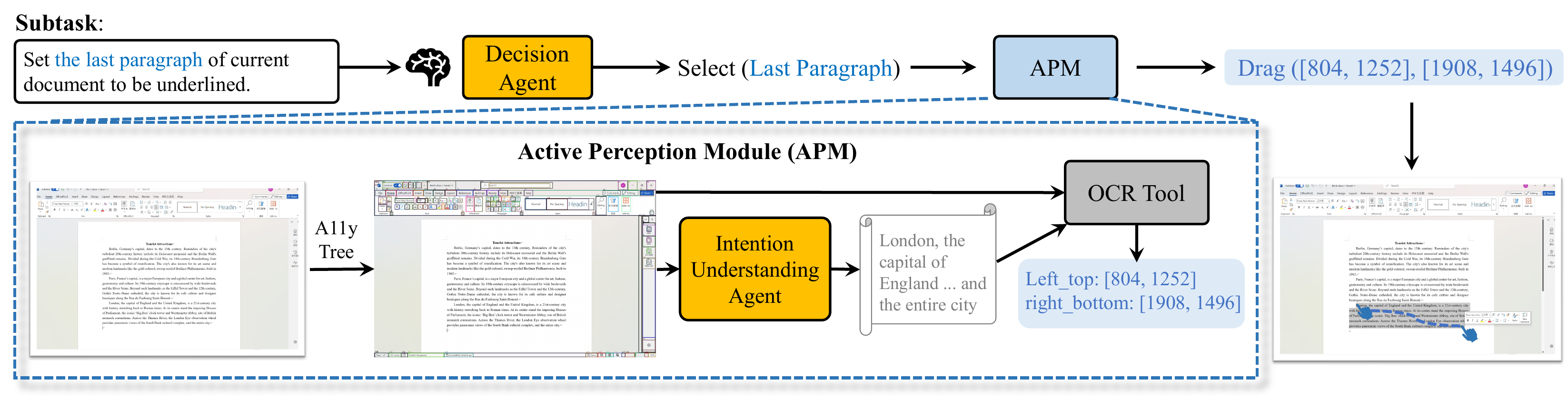}
    \vspace{-2mm}
    \caption{Illustration of the active perception module.
    For interactive elements, the A11y tree is adopted to obtain the bounding boxes and functional information.
    For text, an intention understanding agent and an OCR tool are utilized to perform precise selecting or editing.}
    \label{fig3}
    \vspace{-4mm}
\end{figure*}

\subsection{Active Perception Module}
Firstly, to enable refined perception and operation of interactive elements and text, we propose an active perception module (APM).

\vspace{0.2em}
\noindent\textbf{Interactive element perception.} We first use the \textit{pywinauto} API to extract the accessibility (A11y) tree of the screen interface, filtering and parsing the coordinates and descriptions of the interactive elements. Then we annotate the elements' bounding boxes on the screenshot
in an SoM \cite{yang2023set} manner 
to help MLLM understand the positions and meanings of the elements.

\vspace{0.2em}
\noindent\textbf{Text perception.} Text information cannot be obtained via the A11y tree, and user instructions often reference text vaguely, making it difficult to directly acquire the target text's content and position. For instance, \textit{bold the last two paragraphs of this document}. 
To address this, 
we propose utilizing active perception to obtain the content and position of the target text. As Figure~\ref{fig3} shows, for tasks involving refined text operations (\textit{e.g.}, \textit{selection} or \textit{editing}), 
the decision agent outputs the \textit{Select (target text)} action.
Then the APM employs an MLLM-driven intention understanding agent to determine the start and end range of the target text, followed by using OCR tools to precisely locate the target text for subsequent detailed operations such as \textit{drag}.
A detailed case is shown in Figure~\ref{fig4_2} in the appendix.

\subsection{Hierarchical Multi-agent Collaboration}
\label{sec:hierarchical}

PC scenarios often involve intra- and inter-app workflows, increasing the complexity of user instructions. To address this issue, we employ a divide-and-conquer approach, breaking down the decision-making process into three levels: Instruction, Subtask and Action. As shown in Figure \ref{fig2}, based on this top-down hierarchical decomposition, we design a multi-agent collaboration architecture:
\textbf{(1) Instruction-level:} A manager agent (MA) is set up for high-level task management, which includes decomposing instructions into subtasks, communication between subtasks, and overall progress.
\textbf{(2) Subtask-level:} A progress agent (PA) is established to manage the progress of subtasks.
\textbf{(3) Action-level:} A decision agent (DA) is designated to complete subtasks. Given a specific subtask, the DA makes decisions for each step iteratively, based on the perception of the environment and the operation history provided by the PA.

Through this hierarchical multi-agent collaboration, complex user instructions are decomposed into several interdependent subtasks. The collaborative efforts of manager, progress and decision agents effectively reduce the overall decision-making difficulty and improve the success rate.

\subsubsection{Manager Agent}
In our hierarchical multi-agent collaboration architecture, the LLM-driven manager agent (MA) plays a crucial role in high-level task management:

\noindent\textbf{(1) Instruction decomposition.} As illustrated in Figure~\ref{fig2}, given a complex user instruction, the MA first decomposes it into a series of parameterized subtasks. Each subtask, once instantiated, can be independently executed by the progress agent and decision agent, thus effectively reducing the complexity of individual tasks.

\noindent\textbf{(2) communication between subtasks.} 
The decomposed subtasks often have complex interdependency. Specifically, there are four types of subtasks:
(a) The execution result of the subtask can be used to instantiate subsequent subtasks (\textit{e.g.}, Subtask 1 in Figure~\ref{fig2});
(b) The subtask depends on the execution results of preceding subtasks for instantiation (\textit{e.g.}, Subtask 3 in Figure~\ref{fig2});
(c) The subtask both depends on preceding subtasks for instantiation and produces execution results for subsequent subtasks (\textit{e.g.}, Subtask 2 in Figure~\ref{fig2});
(d) The subtask is independent of other subtasks (\textit{e.g.}, \textit{set an alarm at 10am in the Clock app}).
As Figure~\ref{fig2} shows, during the whole process, the manager agent manages communication between subtasks and complex parameter transmission relationships.
It maintains a communication hub, updates the output of successfully executed subtasks into this hub, and uses the hub to instantiate subsequent subtasks.

\subsubsection{Progress Agent}
After the Manager Agent completes instruction decomposition and necessary inter-subtask communication to instantiate parameterized subtasks, the current independently executable subtask is handed over to the Progress Agent (PA). The PA, also driven by LLM, is responsible for tracking and summarizing the progress of subtasks based on the decisions of the decision agent and the feedback from the reflection agent (will be introduced in Section \ref{sec:reflection}). Once the current subtask is completed, the PA feeds back the output results to the MA.

The purpose of setting up an independent PA between the MA and DA is twofold:
\textbf{(1)} \textbf{It achieves more precise progress tracking} by divide-and-conquer. PA tracks the progress of each subtask individually. This avoids summarizing the entire instruction-level history, which can be lengthy and cumbersome.
\textbf{(2)} \textbf{It facilitates decision-making} by providing the decision agent with a clearer understanding of the operation history and which parts of the subtask remain incomplete. This avoids interference from lengthy history information in the decision-making process.

Specifically, the input for the PA at the $i$-th step includes four parts: (1) the current subtask \(\mathcal{T}\) assigned by the MA; (2) the previous task progress \( \mathcal{TP}_{i-1} \); (3) the action \( \mathcal{A}_i \) output by the $i$-th step's DA; and (4) the reflection \( \mathcal{R}_i \) after executing the $i$-th step's action. Based on this information, the PA outputs the updated progress \( \mathcal{TP}_i \). The above process can be formalized as:

\setlength{\abovedisplayskip}{1pt}
\setlength{\belowdisplayskip}{3pt}
\begin{equation}
\mathcal{TP}_i = PA(\mathcal{T}, \mathcal{TP}_{i-1}, \mathcal{A}_i, \mathcal{R}_i).
\end{equation}

\subsubsection{Decision Agent}
Driven by MLLM, the Decision Agent (DA) is the core agent within the entire PC-Agent framework that generates action decisions and directly interacts with the environment. Given a subtask \(\mathcal{T}\), at each step, DA first obtains an observation \( \mathcal{O}_i \) of the current environment using the perception module. It then combines this with the progress information \( \mathcal{TP}_{i-1} \) output by PA in the previous step, and the reflection information \( \mathcal{R}_{i-1} \) output by RA, to generate the decision for the current step $\mathcal{A}_i$. This process can be formalized as: 

\begin{equation}
\mathcal{A}_i = DA(\mathcal{T}, \mathcal{O}_i, \mathcal{TP}_{i-1}, \mathcal{R}_{i-1}).
\end{equation}

Here, decisions are generated in a Chain-of-Thought~\cite{wei2022chain} manner. An inner monologue for the current step is first generated, followed by the corresponding action decision. This approach not only aids the MLLM in making better decisions but also helps the RA to judge whether the execution results meet expectations.

{After obtaining the decision for the current step, we convert the decision information into a specific action type and corresponding parameters, and then use the \textit{pyautogui} api to execute the corresponding keyboard and mouse operations.
To simplify operations and make decisions easy to parse, we define a constrained action space, which includes \textit{click, double click, type, select, drag, scroll, shortcut} and \textit{stop} (detailed in Appendix~\ref{app_1}).}
This constrained action space ensures that the DA can effectively generate and execute decisions, leading to efficient and accurate task completion.

\subsection{Reflection-based Dynamic Decision-making}
\label{sec:reflection}

Due to factors such as hallucinations and limited reasoning capabilities, even the most advanced MLLMs (\textit{e.g.}, GPT-4o~\citealp{hurst2024gpt}, claude-3.5~\citealp{claude}) find it challenging to avoid errors in perception and decision-making. This issue becomes more pronounced with long operation sequences required by tasks, as a single error in any step can lead to the failure of the entire task.

To detect potential errors in execution results and provide timely feedback and adjustments, we design a reflection-based dynamic decision-making mechanism. Built on the hierarchical architecture introduced in Section \ref{sec:hierarchical}, the dynamic decision-making mechanism operates in a bottom-up manner with the Reflection Agent at its core.

\subsubsection{Reflection Agent}
In the action-level of the hierarchical architecture, we set up a reflection agent (RA) parallel to the decision agent (DA). After the DA makes a decision and executes the corresponding action, the RA observes the change in the system's state before and after the action to determine whether the outcome of this step meets expectations. This process can be formalized as:

\begin{equation}
\mathcal{R}_i = RA(\mathcal{T}, \mathcal{A}_i, \mathcal{O}_{i-1}, \mathcal{O}_i).
\end{equation}

Depending on the execution results, the RA makes three types of judgments:
(1) The execution of the action resulted in changes to the screenshot that did not meet expectations. This may be due to incorrect action type or position parameters in DA's decision, requiring replanning to correct the mistake.
(2) No effective response was produced on the screenshot after executing the action. This might be because the action was executed on a position with no interactive elements, or the element (such as an input box) was not yet activated, necessitating an adjustment in the action execution position.
(3) The action execution produced the correct result, allowing the DA to proceed with the next decision based on this.

In the first two scenarios, the RA's output will be fed back to the DA, enabling the DA to produce decisions based on reflection information to correct errors or avoid repeating ineffective actions. The RA's reflection information will also be fed back to the progress agent (PA), allowing the PA to detect errors and achieve more accurate progress tracking.

\section{Experiments}

\subsection{PC-Eval}

\begin{table*}[t]
\caption{Examples of complex instructions in PC-Eval.}
\label{pc_eval}
\vspace{-1mm}
\centering
\footnotesize
\centering
\begin{tabular}{@{}p{2cm}|p{12cm}|p{1cm}@{}}
\toprule
 \textbf{Applications} & \textbf{Instruction} & \textbf{Steps}\\ 
\midrule
File Explorer \newline Notepad, Clock \newline Calculator & In the Notepad app, open the 'travel\_plan' file in 'Documents', and check the time and location of the travel plans. Add the travel destination to the World Clock list on the Clock app. Calculate the interval between February 18 and the start time of the travel on the Calculator. & 20\\
\midrule
Chrome \newline Excel  &Search on Chrome for the total population of China, the United States, and India in 2024 respectively. Create a new spreadsheet in Excel, write the three countries' names in column A in descending order of population, and the corresponding populations in column B. & 23 \\
\midrule
File Explorer \newline Word & Open the 'test\_doc1' file in 'Documents' in File Explorer, set the title to be bold, and set the line spacing of the first two paragraphs to 1.5x in Word. & 8\\
\bottomrule
\end{tabular}
\end{table*}

\begin{table*}[t]
\caption{Dynamic evaluation results on the PC-Eval benchmark.}
\vspace{-1mm}
\centering
\begin{tabular}{@{}l|c| >{\centering\arraybackslash}p{4cm} >{\centering\arraybackslash}p{3cm}@{}}
\toprule
\multirow{1}{*}{Model} & \multirow{1}{*}{Type}     & Subtask SR (\%) $\uparrow$  & Success Rate (\%) $\uparrow$ \\ \midrule
Gemini-2.0             &   \multirow{4}{*}{Single-Agent}  &   35.4\% & 0.0\%       \\
Claude-3.5             &                               &  15.2\%  & 0.0\%      \\
Qwen2.5-VL              &                               &  46.8\%  & 12.0\%    \\ 
GPT-4o           & &  41.8\%  & 8.0\%     \\      
\midrule
UFO~\cite{ufo}         & \multirow{3}{*}{Multi-Agent} & 43.0\% & 12.0\%  \\
Agent-S~\cite{agashe2024agent} &   &   55.7\%  & 24.0\%    \\
\textbf{PC-Agent} (Ours)                &    & \textbf{76.0\%}   &  \textbf{56.0\%}  \\ \bottomrule
\end{tabular}
\vspace{-2mm}
\end{table*}

Existing benchmarks in real computer environments 
(\textit{e.g.}, OSWorld~\citealp{xie2024osworld} and WindowsAgentArena~\citealp{bonatti2024windows}), though with large scale,
contain mostly basic tasks that may not align with practical workflow requirements, such as \textit{Open Paint and draw a red circle}. 
To better evaluate the abilities of agents on complex PC tasks, we propose a new benchmark PC-Eval, which consists of 25 complex instructions (with 79 subtasks in total) involving 8 popular PC applications (\textit{i.e.}, Chrome, Microsoft Word, Microsoft Excel, Notepad, Clock, Calculator, Outlook and File Explorer). Each instruction comprises several interdependent subtasks, and emphasizes refined operations, practical workflows and long-horizon decision-making.
Three annotators created and checked these instructions to ensure they are realistic and challenging.
Table 1 shows three example instructions, with the complete list in Appendix~\ref{app_2}.
Since different subtasks correspond to different pages and success criteria, creating separate scripts for automatic evaluation of each subtask would be prohibitively costly.
Therefore, we employ human evaluation in this study, and we adopt the following two metrics for evaluation:

\noindent \textbf{(1) Success Rate (SR):} The success rate metric refers to the proportion of successfully completed instructions by the agents.
\noindent \textbf{(2) Subtask Success Rate (SSR):} To comprehensively evaluate the ability of agents, we annotated the subtasks of the PC-Eval instructions, and calculate the success rate of the subtasks completed by the agents.

\subsection{Results}
\textbf{Experimental setup.}
In the experiments, unless otherwise specified, we use GPT-4o as the foundation model for the manager, progress, decision and reflection agents of our PC-Agent framework. 
And we use the OpenOCR \footnote{https://github.com/Topdu/OpenOCR} tool for OCR in the APM.
We compare our PC-Agent with a wide range of single- and multi-agent methods, including advanced MLLMs such as GPT-4o~\cite{hurst2024gpt}, Gemini-2.0~\cite{team2023gemini}, Claude-3.5~\cite{claude}, Qwen2.5-VL 72B~\cite{qwen25}, as well as previous open-source PC agent methods such as UFO~\cite{ufo} and Agent-S~\cite{agashe2024agent}.
For as fair a comparison as possible, we set the same action space for MLLMs via prompting, enabling them to operate as a single decision agent.
As for multi-agent methods UFO and Agent-S, we also adopt GPT-4o as their foundation models.

\begin{table*}[t]
\centering
\caption{The results of the ablation study on the APM module, Manager agent and Reflection Agent.}
\vspace{-1mm}
\begin{tabular}{@{}ccc|cc@{}}
\toprule
\multicolumn{3}{c}{Ablation study}     & \multirow{2}{*}{Subtask Success Rate} & \multirow{2}{*}{Success Rate} \\ \cmidrule(r){1-3}
APM & Manager Agent & Reflection Agent &   & \\ \midrule
    & \checkmark   & \checkmark  & 58.2\% & 20.0\% \\
\checkmark    &     & \checkmark & 50.6\% & 12.0\%  \\
\checkmark    & \checkmark     & &  48.1\% & 12.0\% \\ \midrule
\checkmark    & \checkmark     & \checkmark    &  76.0\%    &   56.0\% \\ \bottomrule
\end{tabular}
\label{tab:ablation}
\end{table*}

\vspace{0.1em}
\noindent \textbf{Results of single agents.}
Table 2 presents the performance comparison of PC-Agent against other methods on PC-Eval. 
It can be seen that those MLLM-based single agents have almost failed on all the instructions.
Even the best-performing Qwen2.5-VL achieves merely a 12\% success rate. This result indicates that relying solely on the abilities of a single decision agent to fulfill complex user instructions on PC is extremely challenging for the current MLLMs.
Meanwhile, the success rate of these models is significantly lower than the subtask success rate. This demonstrates that, due to the lengthy operation sequences and complex dependencies between subtasks, completing the entire instruction is far more difficult than completing the individual subtasks.

\begin{figure*}[t]
    \centering
    \includegraphics[width=0.9 \textwidth]{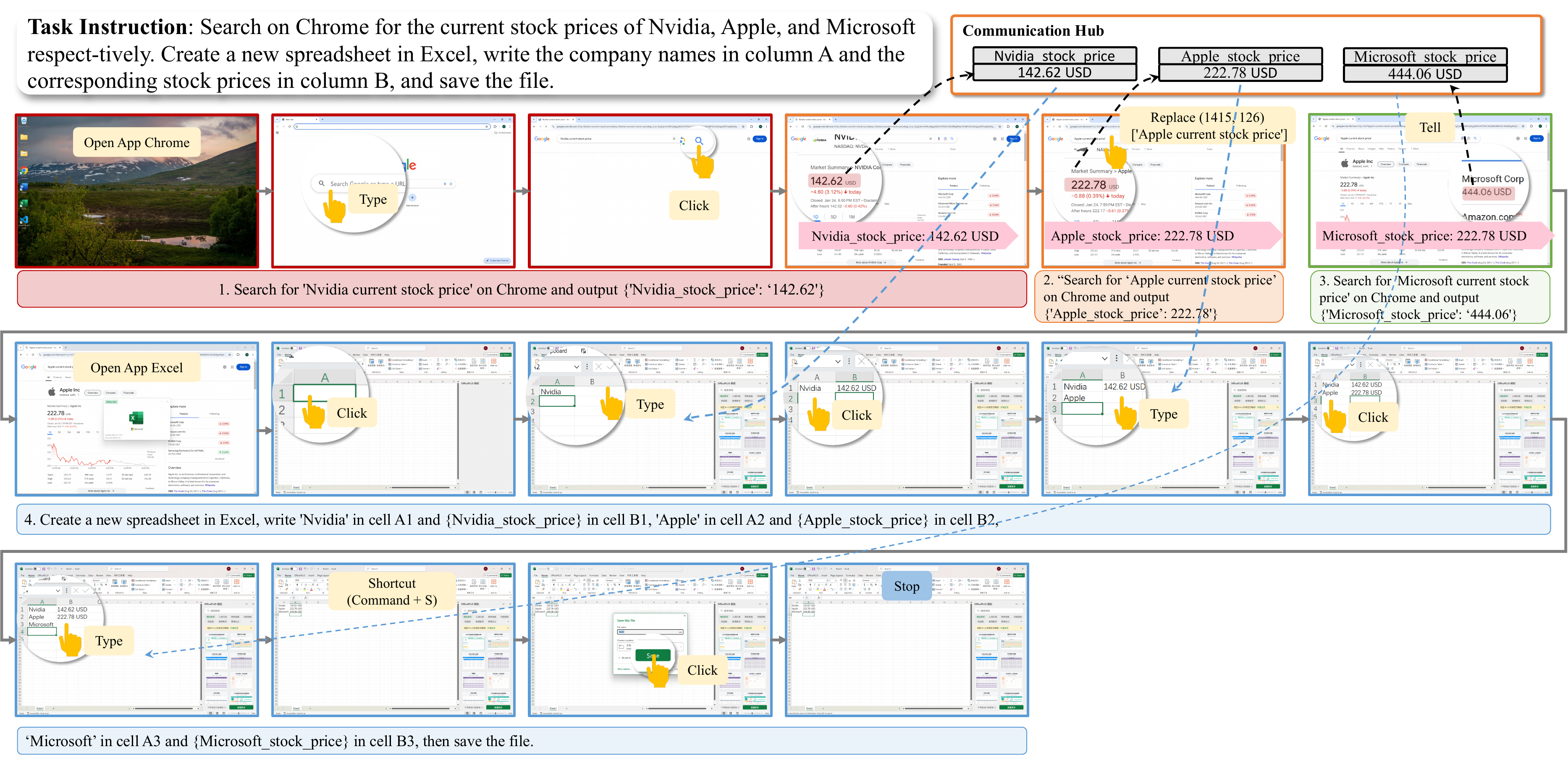}
    \caption{A case of searching for information multiple times and build an Excel sheet accordingly.}
    \label{fig4}
\end{figure*}

\vspace{0.1em}
\noindent \textbf{Results of multi-agent methods.}
UFO and Agent-S are two agent frameworks tailored for PC scenarios. However, on PC-Eval, UFO only achieves a slight advantage over the single agent using GPT-4o. While Agent-S shows an improvement in SSR over single agents, its instruction-level SR remains low. A detailed analysis reveals their problems in both perception and decision-making:

\noindent\textbf{(1)} Existing methods have limited fine-grained perception and operation abilities. For instance, in Excel scenarios such as the one shown in Figure \ref{fig4}, UFO may input multiple pieces of information into the same cell. In Word scenarios such as the one shown in Figure \ref{fig4_2}, both UFO and Agent-S are unable to perform editing operations (\textit{e.g.}, \textit{``underline the last paragraph''}).
\textbf{(2)} Existing methods are insufficient in handling the dependency between subtasks in complex instructions, especially in scenarios where the execution of later subtasks depends on the results of earlier ones. For example, in the instruction ``\textit{... and write down the translation of the content}'', Agent-S would directly write down the text \textit{``The translation of the content''}, rather than the translated content obtained earlier. 

In contrast, our proposed APM enables the PC-Agent to have refined operation abilities. Additionally, through hierarchical multi-agent collaboration, PC-Agent achieves effective instruction decomposition, inter-subtask communication, progress management, and error reflection, which significantly improves the performance on complex tasks. 
As a result, our PC-Agent largely outperforms all previous methods, surpassing UFO and Agent-S by 44\% and 32\% respectively in terms of SR.

\subsection{Ablation Study}

Table \ref{tab:ablation} shows the results of the ablation study on different components of the PC-Agent framework, from which we can conclude:

\noindent \textbf{(1) The active perception module has a significant impact on performance.}
Comparing the first and fourth lines, it can be seen that after removing APM, the SSR decreases by nearly 20\%, while the SR decreases drastically by over 30\%.
On the one hand, without APM, the Decision Agent is unable to grasp the meaning of interactive elements and thus makes more errors. On the other hand, the PC-Agent loses the ability to precisely perceive and manipulate the referred text. As a result, the instruction completion rate has significantly declined.

\noindent \textbf{(2) The manager agent effectively improves PC-Agent's abilities in complex workflow scenarios.}
Comparing the second and fourth lines, it can be seen that removing MA causes SR to significantly decline to 12\%.
This is because without MA, a complex instruction will be treated as a single task for PA and DA to execute. The lengthy operation sequences and complex dependency between the subtasks pose great challenges to progress tracking and also interfere with DA's decision-making.

\noindent \textbf{(3) The reflection-based dynamic decision-making mechanism helps the model recover from errors.}
Comparing the third and fourth lines, it can be seen that removing RA leads to a very significant performance decrease (\textit{i.e.}, 27.9\% in SSR and 44.0\% in SR). This is because during the execution of complex instructions, errors in perception and decision-making are inevitable. Removing RA causes the model to lack awareness and timely correction of errors, 
which predisposes it to getting stuck in meaningless repetition or incorrect steps.

We also conduct ablation experiments on foundation models, please see Appendix ~\ref{app:ablation} for details.

\subsection{Case Study}
Figure~\ref{fig4} illustrates a complete operation process of our PC-Agent framework.
Given a complex user instruction, the Manager Agent first breaks it down into four subtasks. For the first three subtasks, when each is successfully executed, 
the corresponding search result is updated in the communication hub.
Then the MA uses the hub to instantiate the fourth subtask, which reduces the difficulty of the long horizon decision-making process.
Besides, the precise click and type operations in Excel demonstrate the effectiveness of our proposed APM in perceiving complex screen elements.
We also provide a case study on reflection-based dynamic decision-making.
Please refer to Appendix~\ref{app:case} for details.

\section{Related Work}
Recent advances in MLLMs \cite{hurst2024gpt,liu2024visual,wang2024qwen2} have 
inspired research to extend these models to intelligent agents in various domains.
Among these, there's significant focus on GUI Agents for task automation on smart devices.
Currently, research in this field is more concentrated on the Mobile~\cite{zhang2023appagent,wang2024mobilev1,hong2024cogagent} and Web~\cite{gur2023real,zheng2024gpt} scenarios. In the PC scenario, Cradle~\cite{tan2024cradle} focuses on employing MLLM's reasoning abilities to realize operations in AAA games, while PC Agent~\cite{he2024pc} aims to enable agents to create and modify PowerPoint presentations.
Despite the notable progress, their versatility remains relatively limited.
To handle cross-app tasks, UFO~\cite{ufo} designs a dual-agent framework, where one agent is responsible for application selection, and the other agent handles the specific control interactions.
To inject PC task knowledge into decision-making, 
Agent-S~\cite{agashe2024agent} combines online search and local memory for experience-augmented planning.
Compared to previous methods, our PC-Agent focuses on complex PC tasks.
We achieve more refined perception and operation (\textit{e.g.}, editing Word documents) via the devised APM.
And the proposed hierarchical framework realizes a divide-and-conquer pipeline for complex instructions, which effectively addresses the inter-subtask dependencies and significantly improves performance on complex tasks.

\section{Conclusion}
In this work, we proposed a PC-Agent framework to handle complex interactive environments and tasks in PC scenarios.
An Active Perception Module was devised for refined perception and operation capabilities. 
And we proposed a hierarchical multi-agent collaboration architecture to decompose the decision-making process into three levels,
and adopted reflection-based dynamic decision-making for timely error feedback and adjustments.
We created a PC-Eval benchmark of realistic and complex user instructions. Experimental results verify that PC-Agent exhibits superior performance over previous methods on complex PC tasks.

\section*{Limitations}
In this paper, we explored a variety of MLLMs as foundation models. Currently, the best performing model remains the closed-source GPT-4o. However, there is still significant room for improving the efficiency of completing complex tasks by invoking closed-source models.
And the privacy and security issues associated with closed-source models also deserve attention.
Additionally, our focus in this work has primarily been on productivity scenarios. In future work, we will explore expanding into more scenarios such as social interaction and entertainment.

\bibliography{custom}

\appendix

\section{Appendix}
\label{sec:appendix}

\begin{table*}[t]
\centering
\caption{Performance results of PC-Agent with different foundation models on PC-Eval.}
\label{fundation}
\begin{tabular}{@{}lcccc@{}}
\toprule
Model      & Subtask SR (\%) $\uparrow$  & Success Rate (\%) $\uparrow$  & Recovery Rate (\%) $\uparrow$  & Manager SR (\%) $\uparrow$ \\ \midrule
Gemini-2.0 &       55.7\%               &    28.0\%          &   24.0\%  & 84.0\%          \\
Claude-3.5 &     63.3\%          &   40.0\%   &    48.0\%  & 88.0\%  \\

Qwen2.5-VL  &     32.9\%          &   12.0\%     &    40.0\%  &  88.0\%  \\
\midrule
\textbf{GPT-4o}     &     \textbf{76.0\%}          &   \textbf{56.0\%}      &   \textbf{64.0\%}    & \textbf{96.0\%}  \\ \bottomrule
\end{tabular}
\end{table*}

\subsection{Ablation on Foundation Models}
\label{app:ablation}

Table~\ref{fundation} compares the performance of different MLLMs.
Here we introduce two metrics besides SR and SSR to compare the results of using different MLLMs as the foundation model:
\begin{itemize}
    \item \textbf{Recovery Rate:} It measures the proportion of instructions where recovery occurred. The recovery behavior indicates that the agent detects an error and corrects it via reflection (no matter whether the instruction is ultimately completed).
    \item \textbf{Manager SR:} It assesses the ability of the Manager Agent to correctly decompose the user instructions.
\end{itemize}

From the table, we can observe that both the SSR and SR of the PC-Agent driven by GPT-4o are significantly better than the results using Gemini-2.0, Claude-3.5 and Qwen2.5-VL.
And GPT-4o leads Gemini-2.0 by 40\% in terms of the Recovery Rate.
It may be benefit from the better perception and reasoning abilities of GPT-4o.
Besides, it is worth noting that, compared to the single agent using Qwen2.5-VL, the SSR and SR of the PC-Agent using Qwen2.5-VL actually decreased.
Detailed analysis reveals that this is due to Qwen2.5-VL's limited textual ability to follow the format of the output action and unsatisfactory ability to judge whether the task is completed.
And the latter issue becomes more severe after the instruction is decomposed into subtasks.
In conclusion, the result in Table~\ref{fundation} highlights that the abilities of MLLMs are the foundation of the framework's effectiveness.

\subsection{More Case Study}
\label{app:case}

\begin{figure*}[t]
    \centering
    \includegraphics[width=1.0\textwidth]{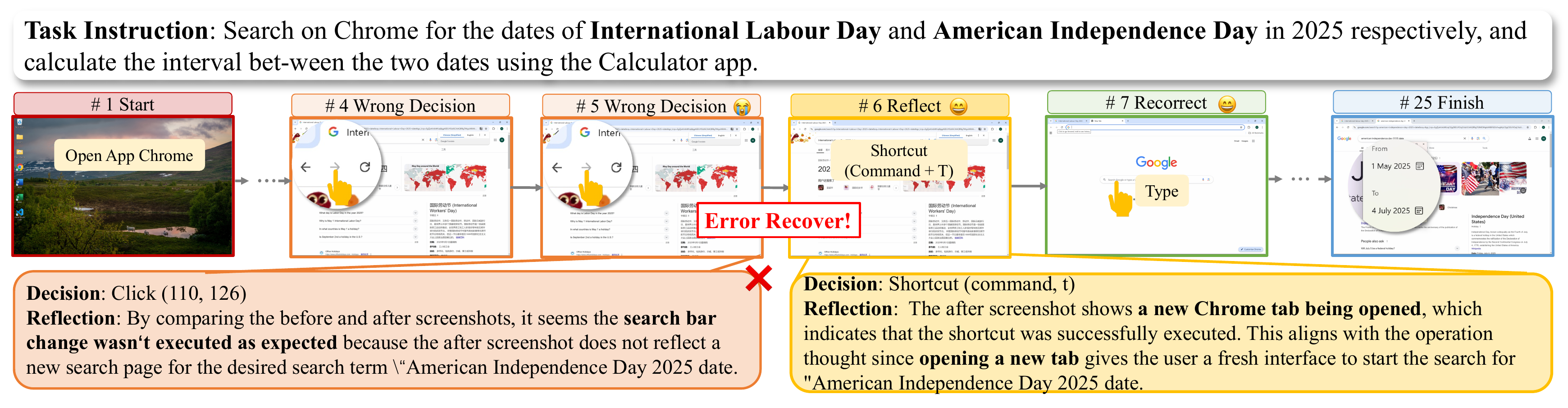}
    \caption{A case of reflection when performing multiple successive searches in Chrome.}
    \label{fig5}
\end{figure*}

Figure~\ref{fig5} shows an example within the PC-Agent framework where the proposed reflection mechanism prevents repetitive invalid operations.
As can be seen,
after the Decision Agent (DA) clicked the \textit{forward button} of the Chrome browser without producing a valid response, the Reflection Agent (RA) detected this error and fed it back to the DA. Based on this feedback, the DA reconsidered in the next step and executed the correct operation (\textit{i.e., use the Shortcut Command + T to open a new tab}).

\begin{figure*}[t]
    \centering
    \includegraphics[width=1.0 \textwidth]{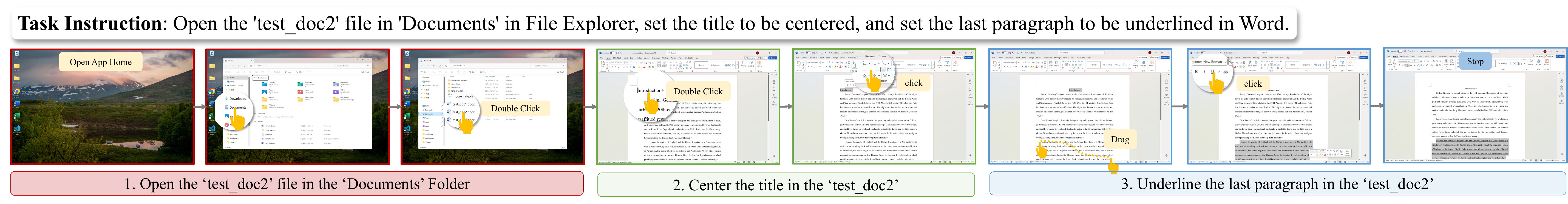}
    \caption{A case of refined text editing operations in the Word application.}
    \label{fig4_2}
\end{figure*}

\subsection{Action Space}
\label{app_1}
We define the action space as follows:
\begin{itemize}
\item Open App (name): Open a specific app using the system's search function.
\item Click (x, y): Click the mouse at position (x, y).
\item Double Click (x, y): Click the mouse twice at position (x, y).
\item {Select (text)}: Acquire the content and position of the target text by invoking the active perception module (APM).
\item Type (x, y) [text]: Input text content at position (x, y).
\item Drag (x1, y1) (x2, y2): Select a specific area of text content by dragging.
\item Scroll (x, y) (value): Scroll the page up or down at position (x, y).
\item Shortcut (key list): Use shortcut keys, such as saving through ctrl+s.
\item Stop: All the requirements have been met, end the current process.
\end{itemize}

\subsection{Instructions in PC-Eval}
\label{app_2}

We show the complete instruction list of PC-Eval as follows:
\begin{itemize}
\item In the Notepad app, open the `memo' file in `Documents', and check the second event in the morning. Set an alarm 1 hour before this event in the Clock app. 
\item  In the Notepad app, open the `memo' file in `Documents', and check the location of the meeting with John. Search on Chrome how much time it takes to get from the Empire State Building to this location.
\item  In the Notepad app, open the `memo' file in `Documents', and check the time and location of the meeting with John. Search on Chrome how much time it takes to get from the Empire State Building to this location, and set an appropriate alarm on the Clock app so that I can leave the Empire State Building in time to arrive at the meeting location punctually.
\item  In the Notepad app, open the `travel\_plan' file in `Documents', and check the travel destination. Use Chrome to search if the traffic at the destination drives on the left or the right.
\item  Search on Chrome for the dates of International Labour Day and American Independence Day in 2025 respectively, and calculate the interval between the two dates using the Calculator app.
\item  Open the `travel\_plan2' file in `Documents' in the Notepad app, and check the three candidate destinations for the travel plan. Search on Chrome for the flight time from Beijing to each destination, and tell me which candidate destination has the shortest flight time.
\item  Search on Chrome for the current stock prices of Nvidia, Apple, and Microsoft respectively. Create a new spreadsheet in Excel, write the company names in column A and the corresponding stock prices in column B.
\item  Search on Chrome for the total population of China, the United States, and India in 2024 respectively. Create a new spreadsheet in Excel, write the three countries' names in column A in descending order of population, and the corresponding population numbers in column B.
\item  Create a new document in Word. Write down two paragraphs introducing Alibaba and OpenAI respectively. Save the document as `TechCompanies'.
\item  Search for the paper `Attention is all you need' on Chrome, download the paper and record its abstract. Create a new document in Word, write down the abstract of the paper, and save it as `Transformer'.
\item  Search for the ratings of `Interstellar' and `12 Angry Men' on imdb.com on Chrome. Open the `movie\_rate' excel file in `Documents' in File Explorer, and fill in the corresponding movie ratings.
\item  Open the `test\_doc1' file in `Documents' in File Explorer, set the title to be bold, and set the line spacing of the first two paragraphs to 1.5x in Word.
\item  Open the `test\_doc2' file in `Documents' in File Explorer, set the title to be centered, and set the last paragraph to be underlined in Word.
\item  Open the `test\_doc3' file in `Documents' in File Explorer, write down the translation of the content below the main text.
\item  Access https://arxiv.org/ in Chrome, search for papers related to `multimodal agent', and download the first paper.
\item  Read the sent mail `Travel' to Howie in Outlook, record the departure, destination and start date of the journey. Search for a one-way flight on booking.com on Chrome.
\item  Search in Chrome for the IMDb ratings of `Leon: The Professional', `The Shawshank Redemption', and `2001: A Space Odyssey'. Record them in a new .txt file using Notepad, sorted from highest to lowest.
\item  Check the sent mail `Code' to Howie in Outlook, download the attachment `homework.py' and open it in Visual Studio Code. Fix the error in this python code.
\item  Create a new Python file in Visual Studio Code, write a function that takes a list as input and outputs the k-th largest number in the list. Send this code file to Howie via Outlook.
\item  Search for tourist attractions in Tokyo and Kyoto respectively in Chrome, and record the information in a new Word document.
\item  Open the `test\_doc3' file located in 'Documents' in File Explorer, note its Chinese content, create a new Word document, and write down the English translation of the Chinese content from test\_doc3.
\item  Open the `test\_doc1' file located in 'Documents' in File Explorer, increase the font size of the title by one level.
\item   In the Notepad app, open the `travel\_plan' file in 'Documents', and check the time and location of the travel plans. Add the travel destination to the World Clock list on the Clock app. Calculate the interval between February 18 and the start time of the travel on the Calculator. 
\item  Search on Chrome for the total population of China, the United States, and India in 2024 respectively. Create a new spreadsheet in Excel, write the three countries' names in column A in descending order of population, and the corresponding populations in column B. 
\item Open the `test\_doc1' file in `Documents' in File Explorer, set the title to be bold, and set the line spacing of the first two paragraphs to 1.5x in Word. 
\item Compare the prices of Amazon, Walmart, and Best Buy for a new Nintendo Switch console in Chrome, and write the site with the cheapest price and the price on Notepad. 
\item   Read the mail `Travel' in Outlook, record the departure, destination and date of the journey. Search for a round-trip flight on booking.com on Chrome. 
\end{itemize}

\subsection{Our GUI Grounding Dataset}
\label{app_3}

\begin{figure*}[t]
    \centering
    \includegraphics[width=1.0\textwidth]{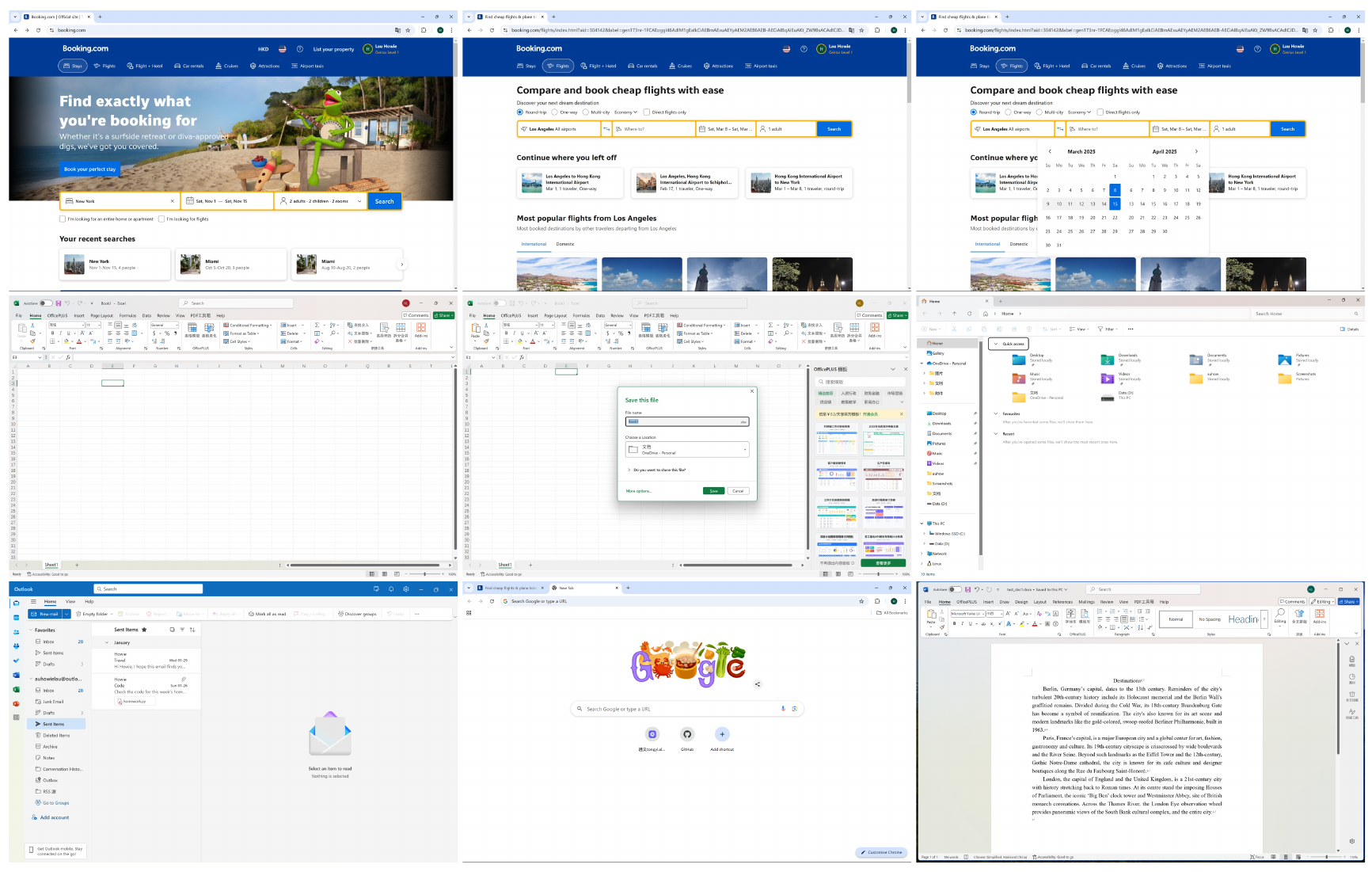}
    \caption{Example screenshots from the GUI grounding dataset we built for commonly used applications in PC scenarios.}
\end{figure*}

On the webpage of Booking.com:
\begin{itemize}

\item Click to book flights
  \item Click to select one-way
  \item Click to select departure location
  \item Click to select destination
  \item Click to select date
  \item Click to select March 21st
  \item Click to select April 1st
  \item Click to select previous month
  \item Click to select next month
\end{itemize}

On the Excel page:
\begin{itemize}
  \item Click to select A3
  \item Click to select E5
  \item Click to select top align
  \item Click to select bottom align
  \item Click to select left align
  \item Click to select right align
  \item Click to save
  \item Click to change file name
  \item Click to change save location
\end{itemize}

On the File Explorer page:
\begin{itemize}
\item Click the Downloads folder
\item Click the Documents folder
\item Click the Pictures folder
\item Click the Music folder
\end{itemize}

On the Outlook page:
\begin{itemize}
\item Click to view inbox
\item Click to view spam/junk email
\item Click to view sent emails
\item Click to view the Travel email sent to Howie
\item Click to view the Code email sent to Howie
\item Click to search
\item Click to create a new email
\item Click to mark as read
\end{itemize}

On the Chrome page:
\begin{itemize}
\item Click the search bar
\item Click the search box
\item Click to open a new tab
\item Click to bookmark
\item Click settings
\item Click refresh
\item Click to switch to Booking.com tab
\end{itemize}

On the Word page:
\begin{itemize}
\item Click for bold
\item Click for italic
\item Click to add underline
\item Click to change text color
\item Click to center text
\item Click to increase font size
\item Click to decrease font size
\item Click to adjust line spacing
\item Click the top-left corner of the title
\item Click the bottom-right corner of the title
\item Click the top-left corner of the second-to-last paragraph
\item Click the bottom-right corner of the second-to-last paragraph
\item Click the bottom-right corner of the last line
\end{itemize}

\end{document}